\documentclass{article}
\usepackage{spconf,amsmath,graphicx}
\usepackage{amsmath,amssymb,graphicx,gensymb}
\usepackage{booktabs}
\usepackage{subcaption} 

\title{Depthwise-STFT based separable Convolutional Neural Networks}
\name{Sudhakar Kumawat, Shanmuganathan Raman}
\address{Computer Science and  Engineering, IIT Gandhinagar}

\name{Sudhakar Kumawat and Shanmuganathan Raman\thanks{S. Kumawat was supported by TCS Research Fellowship and S. Raman was supported by a SERB Core Research Grant.}}
\address{Computer Science and Engineering, Indian Institute of Technology Gandhinagar, India\\\{sudhakar.kumawat, shanmuga\}@iitgn.ac.in}
\begin{document}
%
\maketitle
\begin{abstract}
In this paper, we propose a new convolutional layer called Depthwise-STFT Separable layer that can serve as an alternative to the standard depthwise separable convolutional layer. The construction of the proposed layer is inspired by the fact that the Fourier coefficients can accurately represent important features such as edges in an image. It utilizes the Fourier coefficients computed (channelwise) in the 2D local neighborhood (e.g., $3\times3$) of each position of the input map to obtain the feature maps. The Fourier coefficients are  computed using 2D Short Term Fourier Transform (STFT) at multiple fixed low frequency points in the 2D local neighborhood at each position. These feature maps at different frequency points are then linearly combined using trainable pointwise ($1\times1$) convolutions. We show that the proposed layer outperforms the standard depthwise separable layer based models on the CIFAR-10 and CIFAR-100 image classification datasets with reduced space-time complexity.
		
\end{abstract}
\section{Introduction}
\label{sec:intro}
Over the past few years, with the availability of large-scale datasets and computational power, deep learning has achieved impressive results on a wide range of applications in the field of computer vision. In general, the trend is to achieve higher performance by developing deep and complex models using large computational resources \cite{simonyan2014very, he2016deep, han2017deep, zagoruyko2016wide, szegedy2017inception}. However, this progress is not necessarily making the networks more efficient with respect to memory and speed. In many real world and resource constraint applications such as robotics, satellites, and self-driving cars, the recognition tasks need to be carried out in a fast and computationally efficient manner. Therefore, there is a need to develop space-time efficient models for such applications.

\begin{figure}[t]
	\centering
	\centerline{\includegraphics[width=\columnwidth]{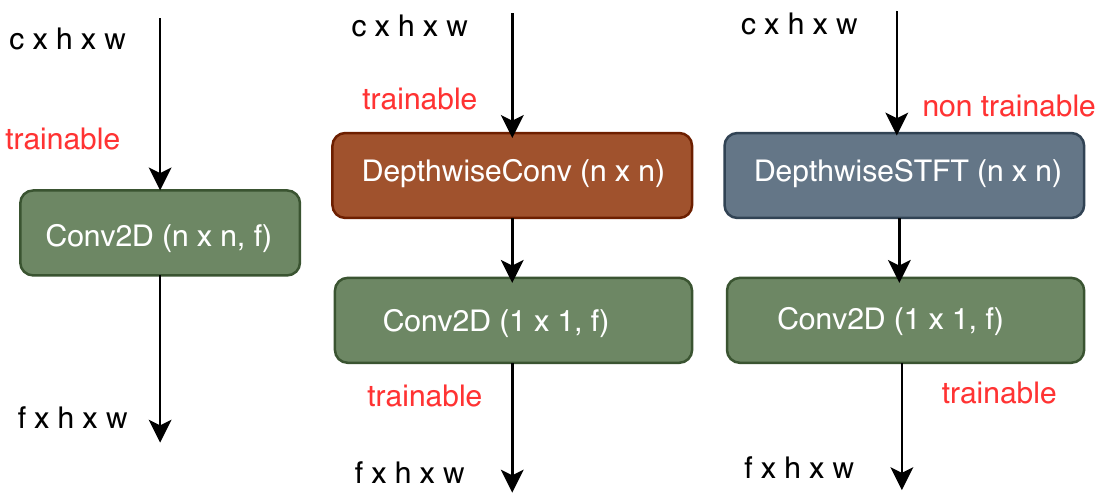}}
	\caption{\textbf{Various convolutional layer architectures.} Standard convolution (left), Depthwise separable convolution (center),  Depthwise-STFT based separable convolution (right). }\label{fig:depthSTFT}
\end{figure}

In a standard convolutional layer, the convolution filters  learn the  spatial and channel correlations simultaneously. The depthwise separable convolutions factorize the above process into two layers. In the first layer, a standard depthwise (channelwise) convolution is applied in order to learn the spatial correlations. In the second layer,  pointwise convolutions ($1\times1$ convolutions) learn channel correlations by combining the outputs of the first layer. Fig.~\ref{fig:depthSTFT}, compares various architectures.  An important advantage of this factorization is that it  significantly reduces the computation and model size. For example, for a filter of size $n=3$, the depthwise separable convolution uses $8$ to $9$ times lesser parameters compared to the standard convolutional layer.

In this paper, we propose a new 2D convolutional layer named Depthwise-STFT separable layer. Similar to the standard depthwise separable convolution layer, our Depthwise-STFT  separable layer  has two sub-layers. The first layer named Depthwise-STFT captures the spatial correlations. For each channel in the input feature map, it computes the Fourier coefficients (at low frequency points) in a 2D local neighborhood ($n\times n$) at each position of the channel to obtain new feature maps. The Fourier coefficients are computed using 2D Short Term Fourier Transform (STFT) at multiple fixed low frequency points in the 2D local neighborhood at each position of the channel. The second layer named pointwise convolution uses  $1\times 1$ convolutions to learn channel correlations by combining feature maps obtained from the Depthwise-STFT layer. Note that unlike the case of standard  depthwise separable layer, here only the second layer (pointwise convolutions) is trainable. Thus, the Depthwise-STFT separable layer has a lower space-time complexity when compared to the depthwise separable convolutions. Furthermore, we show experimentally that the proposed layer achieves better performance compared to the many state-of-the-art depthwise separable based models such as MobileNet \cite{howard2017mobilenets,sandler2018mobilenetv2} and ShuffleNet \cite{ma2018shufflenet,zhang2018shufflenet}.
\begin{figure*}[t]
	\centering
	\centerline{\includegraphics[width=\textwidth, height=0.3\columnwidth]{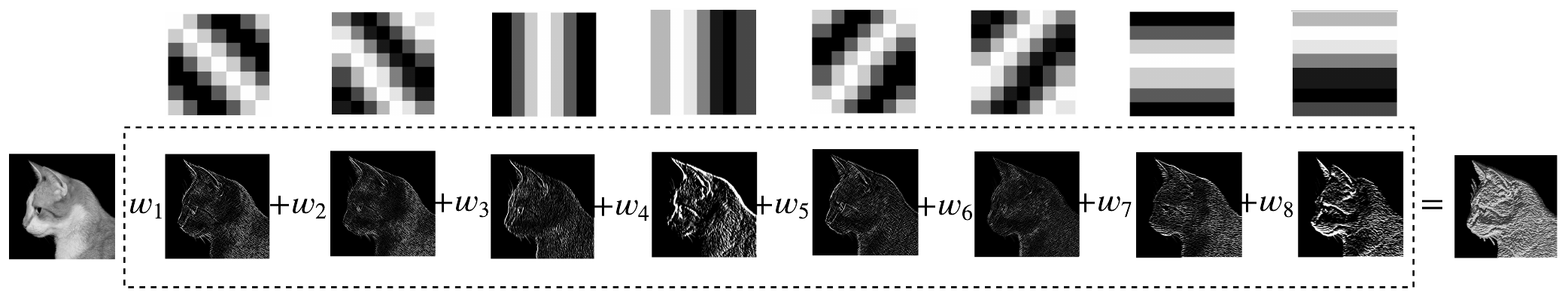}}
	\caption{Visualization of the Depthwise-STFT Separable layer for $c=1$ and $n=7$. Here, weights $w_1, w_2, \dots, w_8$ are learned by the network during training.} \label{fig:billi}
\end{figure*}
\section{Related Works}
\label{sec:related_work}
Recently, there has been a growing interest into developing space-time efficient neural networks for real time and resource restricted applications \cite{mehta2018espnet,howard2017mobilenets,sandler2018mobilenetv2,zhang2018shufflenet,ma2018shufflenet,mehta2019espnetv2,kumawat2019lp}. 

\noindent\textbf{Depthwise Separable Convolutions.} As discussed in Section~\ref{sec:intro}, depthwise separable convolutions significantly reduce the space-time complexity of convolutional neural networks (CNNs) when compared to the standard convolutions by partitioning the steps of learning spatial and channel correlations. Recently, many depthwise separable convolutions based networks have been introduced such as MobileNet \cite{howard2017mobilenets,sandler2018mobilenetv2}, ShuffleNet \cite{ma2018shufflenet,zhang2018shufflenet}, and Xception \cite{chollet2017xception}. Note that with the reduced complexity in the network architectures, these networks achieve trade-off between accuracy and space-time complexity.

\noindent\textbf{2D STFT based Convolutional Layers.} Recently, in \cite{kumawat2019local}, the authors introduced a 2D STFT based 2D convolutional layer named ReLPU for fundus image segmentation. The ReLPU layer when used inplace of the first convolutional layer (following the input layer) of the U-Net architecture \cite{ronneberger2015u} improved the performance of the baseline U-Net architecture.  However, the ReLPU layer could not be used to replace all the convolutional layers in the network due to the extreme bottleneck used in it which reduced its learning capabilities. This work aims to solve this issue by introducing 2D STFT in depthwise separable convolutions.

\section{Method}
\label{sec:method}

We will denote the feature map output by a layer in a 2D CNN network with the tensor $f(\textbf{x})\in \mathbb{R}^{c\times h \times w}$  where $h$, $w$, and $c$ are the height, width, and number of channels of the feature map, respectively. Fig.~\ref{fig:depthSTFT}  presents the high-level architecture of our Depthwise-STFT based separable layer. Note that similar to the standard depthwise separable convolution layer, the Depthwise-STFT based separable layer  has two sub-layers. In the first layer (named Depthwise-STFT), for each channel in the input feature map, we compute the Fourier coefficients in a 2D local neighborhood at each position of the channel to obtain the new feature maps. The Fourier coefficients are computed using 2D Short Term Fourier Transform (STFT) at multiple fixed low frequency points in the 2D local neighborhood at each position of the channel. The second layer (named pointwise convolutions) uses  $1\times 1$ convolutions to learn linear combinations of the feature maps obtained from the Depthwise-STFT layer.  The detailed description of each layer is as follows.\\

\noindent\textbf{Depthwise-STFT.}  This layer takes a feature map $f(\textbf{x}) \in \mathbb{R}^{c\times h \times w}$ as input from the previous layer. For simplicity, let us take $c=1$. Hence, we can drop the channel dimension and rewrite the size of $f(\textbf{x})$ as $  h\times w$. Here, $\textbf{x}\in\mathbb{Z}^2$ are the 2D coordinates of the elements in $f(\textbf{x})$.

Next, each $\textbf{x}$ in $f(\textbf{x})$  has a $n\times n$ 2D neighborhood (denoted by $\mathcal{N}_{\textbf{x}}$) which can be defined as shown in Equation~\ref{eq:1}.
\begin{equation}\label{eq:1}
\mathcal{N}_{\textbf{x}}=\{\textbf{y}\in\mathbb{Z}^2 \,; \parallel(\textbf{x}-\textbf{y})\parallel_{\infty} \leq r \, ; n=2r+1; r\in\mathbb{Z}_{+}\}
\end{equation}
Now, for all positions $\textbf{x}=\{\textbf{x}_1,\textbf{x}_2,\ldots,\textbf{x}_{h\times w}\}$ of the feature map $f(\textbf{x})$, we use local  2D neighborhoods, $ f(\textbf{x}-\textbf{y}) ,\forall \textbf{y}\in \mathcal{N}_{\textbf{x}}$ to derive the local frequency domain representation. For this, we use Short Term Fourier Transform (STFT) which is defined in Equation~\ref{eq:2}.
\begin{equation}\label{eq:2}
F(\textbf{v},\textbf{x})= \sum_{\textbf{y}_i\in \mathcal{N}_\textbf{x}} f(\textbf{x}-\textbf{y}_i)\exp^{-j2\pi \textbf{v}^T \textbf{y}_i}
\end{equation}
Here $i = 1,\ldots,n^2$, $\textbf{v}\in\mathbb{R}^2$ is a 2D frequency variable and $j=\sqrt{-1}$. Note that, due to the separability of the basis functions, 2D STFT can be efficiently computed using simple 1D convolutions for the rows and the columns, successively. Using vector notation, we can rewrite Equation~\ref{eq:2} as shown in Equation~\ref{eq:3}.  
\begin{equation}\label{eq:3}
F(\textbf{v},\textbf{x})=\textbf{w}^T_\textbf{v}\textbf{f}_\textbf{x}
\end{equation}
Here,  $\textbf{w}_\textbf{v}$ is a complex valued basis function (at frequency variable $\textbf{v}$) of a linear transformation and is defined as shown in Equation~\ref{eq:4}.
\begin{equation}\label{eq:4}
\textbf{w}^T_\textbf{v}=[\exp^{-j2\pi\textbf{v}^T\textbf{y}_1}, \exp^{-j2\pi\textbf{v}^T\textbf{y}_2},\ldots, \exp^{-j2\pi\textbf{v}^T\textbf{y}_{n^2}}] 	
\end{equation}
$\textbf{f}_\textbf{x}$ is a vector containing all the elements from the neighborhood $\mathcal{N}_\textbf{x}$ and is defined as shown in Equation~\ref{eq:5}.
\begin{equation}\label{eq:5}
\textbf{f}_\textbf{x}=[f(\textbf{x}-\textbf{y}_1), f(\textbf{x}-\textbf{y}_2),\ldots, f(\textbf{x}-\textbf{y}_{n^2})]^T
\end{equation}
In this work, we use four lowest non-zero frequency variables $\textbf{v}_1= [a,0]^T, \textbf{v}_2=[0,a]^T,\textbf{v}_{3}=[a,a]^T$, and $\textbf{v}_{4}=[a,-a]^T$, where $a=1/n$.  Thus, from Equation~\ref{eq:3}, we can define the local frequency domain representation for the above four frequency variables  as shown in Equation~\ref{eq:6}.
\begin{equation}\label{eq:6}
\textbf{F}_{\textbf{x}} = [F(\textbf{v}_1,\textbf{x}), F(\textbf{v}_2,\textbf{x}),F(\textbf{v}_3,\textbf{x}), F(\textbf{v}_{4},\textbf{x})]^T 
\end{equation}
At each position $\textbf{x}$, after separating the real and the imaginary parts of each component, we obtain a vector as shown in Equation~\ref{eq:7}.
\begin{multline}\label{eq:7}
\textbf{F}_{\textbf{x}} = [\Re\{F(\textbf{v}_1,\textbf{x})\}, \Im\{F(\textbf{v}_1,\textbf{x})\},\ldots\\
\ldots,\Re\{F(\textbf{v}_{4},\textbf{x})\}, \Im\{F(\textbf{v}_{4},\textbf{x})\}]^T
\end{multline}
Here, $\Re\{\cdot\}$ and $\Im\{\cdot\}$  return the real and  imaginary parts of a complex number, respectively. The corresponding $ 8 \times n^2 $ transformation matrix can be written as shown in Equation~\ref{eq:8}.
\begin{equation}\label{eq:8}
\textbf{W} = [\Re\{\textbf{w}_{\textbf{v}_1}\}, \Im\{\textbf{w}_{\textbf{v}_1}\}, \ldots, \Re\{\textbf{w}_{\textbf{v}_{4}}\}, \Im\{\textbf{w}_{\textbf{v}_{4}}\}]^T
\end{equation}
Hence, from Equations~\ref{eq:3} and~\ref{eq:8}, the vector form of STFT for all the four frequency points $\textbf{v}_1, \textbf{v}_2, \textbf{v}_3 $, and $\textbf{v}_{4}$ can be written as shown in Equation~\ref{eq:9}.
\begin{equation}\label{eq:9}
\textbf{F}_\textbf{x}=\textbf{Wf}_\textbf{x}
\end{equation}
Since $\textbf{F}_\textbf{x}$ is computed for all positions $\textbf{x}$ of the input $f(\textbf{x})$, it results in an output feature map with size $8\times h\times w$ corresponding to the four frequency variables. Remember that we took $c=1$. Thus, for one channel, the Depthwise-STFT layer outputs a feature map of size  $8\times h\times w$ corresponding to the four frequency variables. Therefore, for $c$ channels, the Depthwise-STFT will output a feature map of size $8\cdot c\times h\times w$.

\noindent\textbf{Pointwise convolutions.} This layer is the standard trainable $1\times 1$ convolutional layer containing $f$ filters, each one of them has a depth equal to $8\cdot c$ which takes as input a tensor of size $8\cdot c\times h\times w$ and outputs a tensor of size $f\times h\times w$. Note that it is this layer that gets learned during the training phase of the CNN.

In Fig.~\ref{fig:billi}, we present the visualization of the Depthwise-STFT based separable layer for $c=1$ and $n=7$. First the Fourier coefficients of the input feature map is extracted in a local $7\times7$ neighborhood of each position at four frequency points $[1/7, 0]^T, [0, 1/7]^T, [1/7, 1/7]^T,$ and $[1/7, -1/7]$ to output the feature maps of size $8\cdot c\times h\times w$ (after separating real and imaginary parts). Then, the output feature maps are  linearly combined to output the final feature map.


\noindent\textbf{Parameter analysis.} Consider a standard 2D convolutional layer with $c$ input  and $f$ output channels. Assume that spatial padding is done such that the spatial dimensions of the channels remain same. Let $n\times n$ be the size of the filters. In  Table~\ref{tab:params}, we compare the number of trainable parameters in various convolutional layers. The number of trainable parameters in Depthwise-STFT separable layer is independent of the filter size $n$.

\begin{table}[t]
	\centering
	\footnotesize
	\begin{tabular}{lc}  
		\toprule
		\textbf{Layer} &\textbf{\# parameters}  \\
		\midrule
		Standard Convolution		& $c \cdot n^2 \cdot f $  	\\
	
		Depthwise Separable	& $n^2 \cdot c + c \cdot f $	  \\

		Depthwise-STFT Separable	& $8\cdot c\cdot f$	  \\
		
		\bottomrule
	\end{tabular}
	\caption{Comparison of the number of trainable parameters in various convolutional layers.}
	\label{tab:params}
\end{table}

\begin{figure*}[t]
	\begin{subfigure}[b]{\columnwidth}
		\centering
		\includegraphics[width=.6\columnwidth, height=0.6\columnwidth]{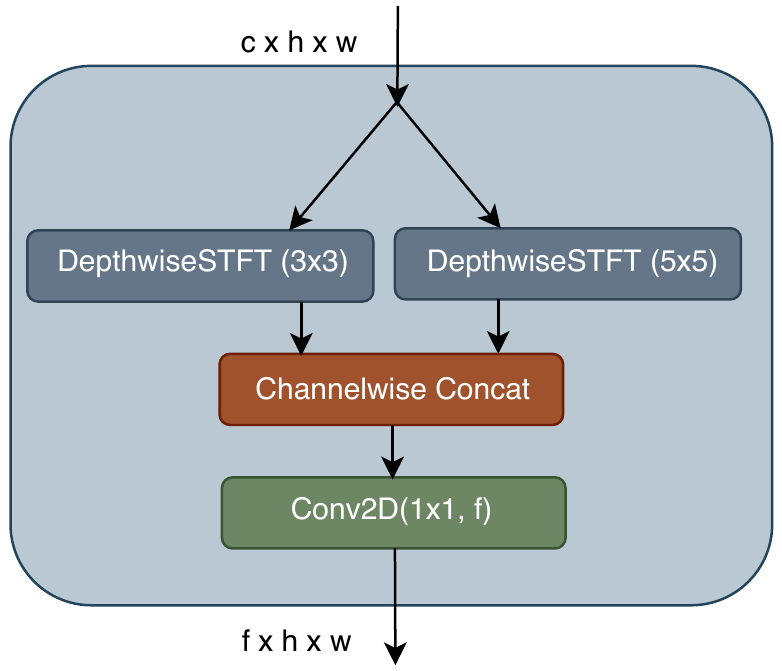}
		\caption{Block 1}
		\label{fig:a}
	\end{subfigure}%
	\begin{subfigure}[b]{\columnwidth}
		\centering
		\includegraphics[width=.82\columnwidth, height=0.6\columnwidth]{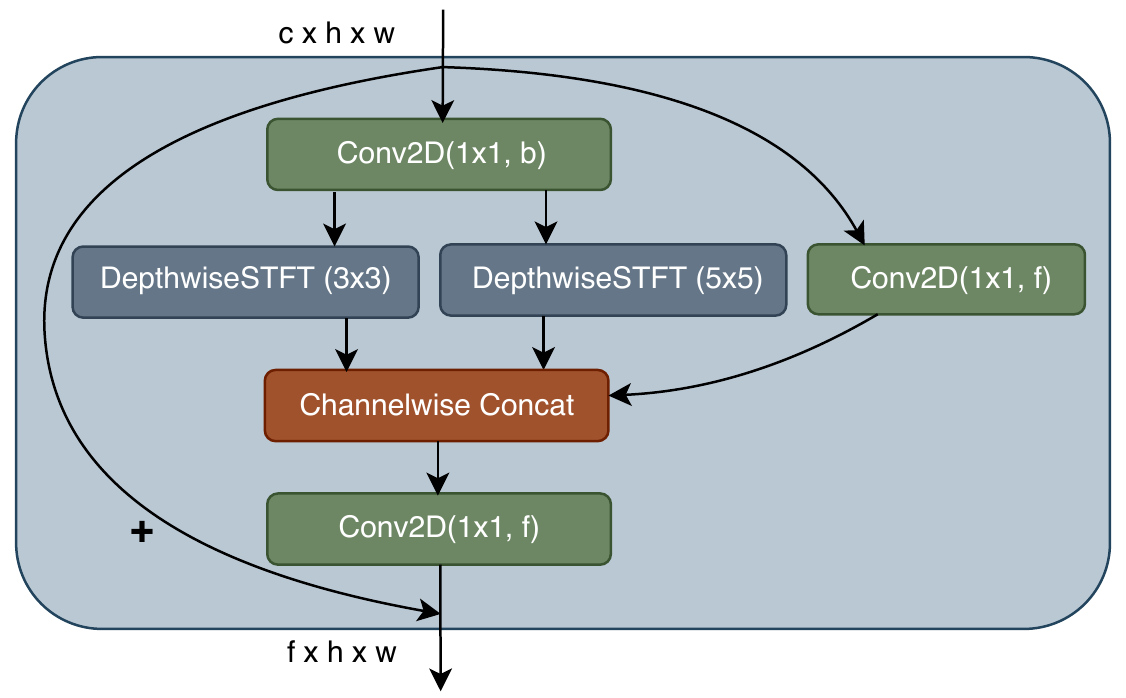}
		\caption{Block 2}
		\label{fig:b}
	\end{subfigure}		
	
	\begin{subfigure}[b]{\textwidth}
		\centering
		\includegraphics[height=0.13\columnwidth]{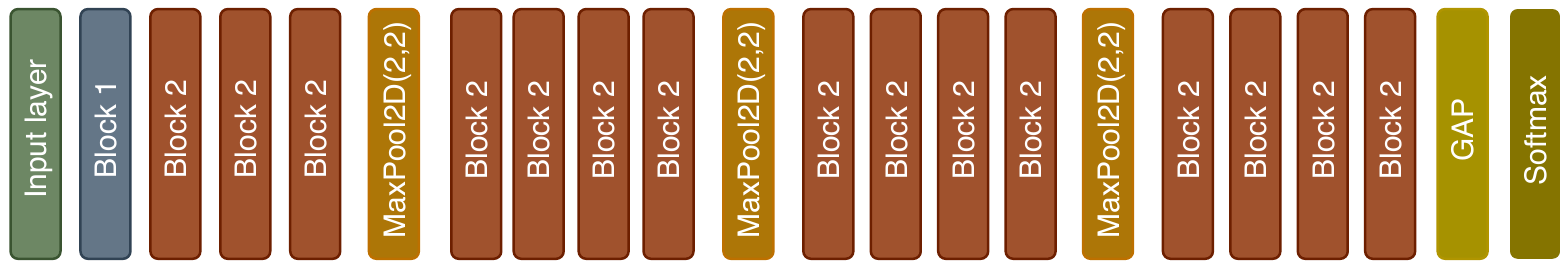}
		\caption{Our proposed network architecture}
		\label{fig:voxception_block}
	\end{subfigure}
	\caption{Various building blocks of the proposed network architecture and our proposed network} \label{fig:blocks}
\end{figure*}

\section{Experiments}
\noindent
\textbf{Datasets.} We evaluate the  Depthwise-STFT separable layer on the two popular  CIFAR datasets$-$ CIFAR-10 and CIFAR-100 \cite{krizhevsky2009learning}. The CIFAR-10 and the CIFAR-100 datasets consist of 10 and 100 classes, respectively. Each  dataset consists of natural RGB images of resolution $32\times32$ pixels  with 50,000 images in the training set and 10,000 images in the testing set. We use the standard data augmentation scheme $-$ horizontal flip/shifting/rotation $(\pm20)$ that is widely used for these two datasets. For preprocessing, we normalize the data using the channel means and standard deviations. 

\noindent\textbf{Network Architecture.}  We adopt a simple Inception-ResNet style bottleneck architecture \cite{szegedy2017inception}. Fig.~\ref{fig:blocks} presents the building blocks of our network, \emph{Block 1} (Fig.~\ref{fig:a}) and \emph{Block 2} (Fig.~\ref{fig:b}). \emph{Block 1} takes as input a feature map with $c$ channels and applies a bottleneck $1\times 1$ convolution (trainable) to output a feature map of size $b$ such that $b<c$. This is followed by inception style non-trainable Depthwise-STFT layers with filter size $3$ and $5$. The intuition behind this design is to have a maximum number of possible pathways for information to flow through the network and let the network selectively choose the best frequency points/neighborhood sizes for computing local Fourier transform and to selectively give more weight to them during training. The outputs from the Depthwise-STFT layers are concatenated channel-wise and finally passed through an expansion $1\times 1$ convolution (trainable) to output a feature map of size $f$ such that $f>b$. We use bottleneck architectures as they have been proved to be generalize better \cite{sandler2018mobilenetv2}. The architecture of \emph{Block 2} is just a slightly augmented version of  \emph{Block 1} with skip connections (Fig.~\ref{fig:b}). Each block is followed by a batch normalization layer which is followed by a LeakyReLU (with $\alpha=0.3$) activation function \cite{maas2013rectifier}. For down-sampling, we use max pooling with pool size 2 and stride 2. Our final network as shown in Fig.~\ref{fig:voxception_block} consists of 16 non-downsampling blocks (\emph{Block 1} and \emph{Block 2}), followed by a global average pooling layer connected to a final classification layer with softmax activation. We implemented the proposed network using Keras deep learning library \cite{gulli2017deep} and performed all the experiments on a system with Intel i7-8700 processor, 32 Gb RAM, and a single Nvidia Titan Xp GPU.

\begin{table}[t]
	\centering
	\footnotesize
	\begin{tabular}{lcccc}  
		\toprule
		\textbf{Network}&\textbf{\# params}&\textbf{\# feat.}&\textbf{CIFAR-10}&\textbf{CIFAR-100} \\
		\midrule
		ShuffleNet \cite{zhang2018shufflenet}		& 1.05M  &	800     &  90.80	&  70.06	\\
		ShuffleNet-V2 \cite{ma2018shufflenet}	& 1.35M	  & 1024	&  91.42    &  69.51	\\
		MobileNet \cite{howard2017mobilenets} 		& 3.31M	  & 1024	&  91.87	&  65.98    \\
		MobileNet-V2 \cite{sandler2018mobilenetv2}	& 2.36M	  & 1280	&  93.14	&  68.08    \\
		ReLPU based \cite{kumawat2019local}	& 3.08M	  & 256	&  92.20	&  70.20    \\
		\midrule
		Ours (b=8, f=128) & 0.90M   & 128	& 93.16	& 70.19 \\
		Ours (b=16, f=128) & 1.30M  & 128	& 93.59	& 70.66\\
		Ours (b=32, f=128) & 2.21M & 128	&  93.72& 71.08 \\
		Ours (b=64, f=128) & 	3.69M	& 128	& 94.07	& 71.42 \\
		\midrule	
		Ours (b=64, f=256) & 8.21M	& 256	& 94.25	& 73.01\\
		Ours (b=64, f=384) & 14.06M	& 384	&  \textbf{94.51} & \textbf{74.39} \\
		\bottomrule
	\end{tabular}
	\caption{Performance results of the Depthwise-STFT separable layer based architectures compared to the standard depthwise separable layer based architectures.}
	\label{tab:results}
\end{table}

\noindent\textbf{Training.} For training out networks, we use Adam optimizer \cite{kingma2014adam}, categorical cross-entropy as loss function, and batch size of 64. All the trainable weights are initialized with orthogonal initializer \cite{he2015delving}. We train our networks with a learning rate of 0.01 for 300 epochs. After that, we increase the batch size to 128 and further train the networks for 100 epochs. Note that the above training method is inspired from \cite{smith2017don} which proposes that it is preferable to increase the batch size instead of decreasing the learning rate during training.

\noindent\textbf{Results and Analysis.} Table~\ref{tab:results} reports the classification results of different resource efficient architectures  on the CIFAR-10 and CIFAR-100 datasets. For fair comparison, we compare our networks with depthwise separable based architectures only such as MobileNet \cite{howard2017mobilenets,sandler2018mobilenetv2} and ShuffleNet \cite{ma2018shufflenet,zhang2018shufflenet}. As discussed in Section~\ref{sec:related_work}, we also compare with the ReLPU layer \cite{kumawat2019local} based network which is form by replacing all the layers of the network of Fig.~\ref{fig:voxception_block} with ReLPU layer. Our results show that the proposed layer  outperforms the depthwise separable layer based and the ReLPU layer based architectures. Note that here our target is not to achieve state-of-the-art accuracy on the two datasets but to showcase the efficiency of our proposed layer when compared to the depthwise separable convolution based architectures. For analysis purpose, we ran two variants of the proposed network (Fig.~\ref{fig:voxception_block}). In the first variant, the bottleneck parameter $b$ is increased while the expansion parameter $f$ is kept constant. In the second variant, we kept the  bottleneck parameter $b$ constant while increasing the expansion parameter $f$. In both the settings, we observe improvement in the performance of the networks.

\section{Conclusion}
This paper proposes Depthwise-STFT separable layer, that can serve as an alternative to the standard depthwise separable layer. The proposed layer captures  spatial correlations (channelwise) in the feature maps using  STFT followed by pointwise convolutions to channel correlations.  Our proposed layer outperforms the standard depthwise separable layer based models on the CIFAR-10 and CIFAR-100  datasets. Furthermore, it has lower space-time complexity when compared to standard depthwise separable layer.

\end{document}